\documentclass[12pt]{article}
\usepackage[margin=1in]{geometry}
\usepackage{setspace}
\usepackage{amsmath,amssymb}
\usepackage{booktabs,longtable,array,tabularx}
\usepackage{ragged2e}
\usepackage{graphicx}
\usepackage{microtype}
\usepackage{enumitem}
\usepackage{caption}
\usepackage{subcaption}
\usepackage{float}
\usepackage{xurl}
\usepackage[hidelinks]{hyperref}
\usepackage[style=apa,backend=biber,sortcites=true,sorting=nyt]{biblatex}
\graphicspath{{figures/}}
\newcommand{\Y}{\mathbf{Y}}

\title{\textbf{Machine Zygote: Causal Biparental Heredity Before Learning in a Germline--Soma Artificial Agent}}
\author{Lyes Saad Saoud\\\small Independent Researcher\\\small Chicago, Illinois, USA; Abu Dhabi, UAE}
\date{}

\begin{document}
\maketitle

\begin{abstract}
Artificial ontogeny, developmental encodings, robot reproduction, and inherited controllers are established research directions, yet they leave open a narrower experimental question: can a newborn artificial agent exhibit measurable biparental heredity before learning, and can that dependence be isolated causally rather than inferred only from parent--offspring resemblance? We introduce \emph{Machine Zygote}, a computational germline--soma architecture designed around that question. Two parental germline vectors are independently mutated and recombined into a zygote; the zygote parameterizes a developmental regulatory system acting on an initially generic eight-module soma; the resulting soma is frozen and evaluated without learning. A preregistered $4\times4$ diallel comprising 640 offspring shows significant dam and sire dependence for five of six newborn behavioral traits after Holm correction. Parental and interaction variance components account for 36--53\% of modeled variance for five principal traits. In matched-background interventions ($n=60$), replacing only one parental germline while holding the recombination mask, mutation vectors, and developmental-noise seed fixed causes phenotype shifts that exceed a same-parent re-mutation control for five of six traits for both parental channels. Recombination also produces excess transgressive offspring for speed and gait frequency. A preregistered developmental-dependence hypothesis is not supported: a quasistatic no-dynamics ablation preserves the mean phenotype distribution while altering its parental variance structure. Thus the present evidence supports causal biparental, pre-learning heredity in this simulation, but not the stronger claim that recurrent developmental dynamics are necessary for that heredity. The study does not establish physical heredity, biological genetics, or autonomous evolution. Its contribution is an intervention-centered framework for studying inherited newborn phenotypes in artificial agents and a reproducible benchmark for separating heredity, development, stochastic variation, and post-birth learning.
\end{abstract}


\section{Introduction}
\begin{figure}[t]
\centering
\includegraphics[width=0.93\textwidth]{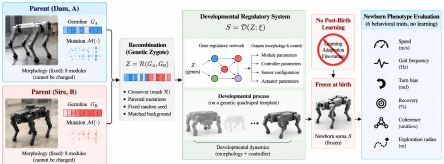}
\caption{Machine Zygote computational architecture. Two parental germline channels are independently perturbed and recombined into a zygote, which parameterizes a developmental regulatory process and produces a newborn operational soma. The soma is frozen before phenotype measurement, and no post-birth learning or adaptation occurs during evaluation. The robotic forms are conceptual illustrations of parental and offspring roles only; the reported experiments use the computational model and differential-drive body abstraction described in the text, not physical quadruped robots.}
\label{fig:schematic}
\end{figure}

Artificial-life and evolutionary-robotics research has long asked how complex agents can be generated rather than manually specified. Developmental encodings allow compact genotypes to unfold into larger phenotypes \parencite{stanley2003,bongard2001}; ontogenetic hardware has translated developmental metaphors into self-replicating, self-repairing, and growing electronic systems \parencite{sipper1997}; evolutionary robotics has linked evolved digital descriptions to automatically fabricated machines \parencite{lipson2000}; and embodied artificial evolution has proposed populations of physical robots that reproduce and evolve in real time and space \parencite{eiben2012,eiben2013}. More recent work has examined inheritance of learned controllers in simulated modular robots \parencite{jelisavcic2019,luo2023}, as well as richer morphogenetic mechanisms for robot design \parencite{hockings2020}. These precedents make one point clear: neither ``zygote,'' developmental encoding, robot reproduction, nor inherited control is itself a new primitive.

The unresolved issue addressed here is more specific. Most claims of artificial heredity are demonstrated by a generative mapping, a parent--offspring correlation, or the propagation of genotype/controller parameters. Those are important demonstrations, but they do not by themselves establish which parental contribution causes which newborn phenotypic change under a matched background. Likewise, many evolving-robot systems include learning immediately after birth, making inherited state and post-birth adaptation difficult to separate conceptually. In a recent Lamarckian robot-evolution system, for example, learned controller weights are encoded back into the genotype and inherited, and newborn fitness is evaluated around a subsequent learning process \parencite{luo2023}. Our question is deliberately different: \emph{before any learning occurs, can two parental germlines jointly determine a newborn behavioral phenotype, and can changing one parental germline alone cause a reproducible shift?}

We operationalize that question through a germline--soma separation. An individual is
\begin{equation}
M=(G,S),
\end{equation}
where $G$ is used only to generate descendants and $S$ is the operational controller/body abstraction. Reproduction generates a zygote $Z$ from two parental germlines,
\begin{equation}
Z=\mathcal{R}(G_{\mathrm{dam}},G_{\mathrm{sire}},\xi),
\end{equation}
and a fixed developmental/readout pipeline produces the newborn soma,
\begin{equation}
S^{(0)}=\mathcal{D}(Z,U,E_0),
\end{equation}
from an identical generic soma $U$ and standardized environment $E_0$. The newborn phenotype $\Y=\Phi(S^{(0)})$ is then measured with the controller held fixed. The architecture is therefore a computational analogue of germline--soma separation, not a claim of biological equivalence and not a physical germline.

The study was built around five preregistered hypotheses. H1 tests whether both parental identities contribute to newborn phenotype in a balanced diallel. H2 tests whether dynamical development is necessary to organize that phenotype. H3 tests whether the H1 effects are expressed before learning by evaluating newborns under a fixed, non-adaptive controller. H4 tests whether recombination creates transgressive offspring beyond clone-line noise. H5 performs the strongest test: a matched-background intervention replaces only one parental germline while holding the recombination and stochastic background fixed.

The principal contribution is not a claim to have invented artificial embryogeny. Instead, it is an \emph{intervention-centered experimental formalization of biparental pre-learning heredity} in an artificial agent. The study combines (i) a balanced $4\times4$ diallel, (ii) an explicit germline--soma separation, (iii) newborn evaluation with no plasticity or learning, (iv) same-background parental-germline interventions, (v) direct-controller, random-germline, and no-development comparisons, and (vi) preregistered decision rules that retain a failed developmental hypothesis rather than rewriting the story around the observed data.

\section{Relation to Existing Work on Artificial Heredity}

\subsection{Developmental encodings and artificial embryogeny}
Developmental genotype--phenotype mappings are established. \textcite{stanley2003} formalized artificial embryogeny as evolutionary systems in which phenotypes undergo a developmental phase, while \textcite{bongard2001} used gene-regulatory development to generate morphology and control in virtual agents. Development can alter both evolutionary search and the relation between genotype and phenotype; for example, \textcite{shreesha2023} compared hardwired and regulative artificial embryogeny and showed that developmental competency changes evolutionary dynamics. Our model inherits that general lineage: a compact hereditary state is transformed into a richer operational state through regulatory dynamics and a fixed differentiation rule.

Accordingly, the present paper does not claim that a germline separated from a developed phenotype is unprecedented. Its narrower contribution is the use of this separation as the basis of a causal heredity experiment: two parent channels, reciprocal crosses, a full factorial diallel, a no-learning newborn endpoint, and matched substitutions of one parent while all stochastic background variables are fixed.

\subsection{Robot reproduction and inherited controllers}
Automated fabrication and self-reproduction have physical precedents \parencite{lipson2000,zykov2005}. The Triangle of Life explicitly treats conception, morphogenesis, and learning as stages of embodied robot evolution \parencite{eiben2013}. Contemporary robot-evolution discussions similarly envisage digital genotypes that are recombined and subsequently mapped into new physical phenotypes \parencite{eiben2021}. Thus a future physical Machine Zygote would enter a mature landscape of embodied evolution rather than an empty field.

Controller heredity is also established. \textcite{jelisavcic2019} inherited parental controller structure to bootstrap newborn learning, and \textcite{luo2023} made learned controller weights inheritable through a reversible genotype--phenotype mapping. Their systems answer questions about learning and evolution. The present study removes learning from the newborn endpoint so that inherited phenotype can be examined before adaptation and then intervened on directly.

\subsection{What is and is not claimed as new}
We use the term \emph{Machine Zygote} as a memorable name for the computational object formed by biparental recombination. Terms based on zygotes and embryogenesis have a substantial prior history in artificial life and hardware \parencite{sipper1997,stanley2003}. Our novelty claim is therefore intentionally modest and falsifiable: the combination of a germline--soma architecture with a balanced biparental design and matched germline interventions provides a reproducible method for testing causal, pre-learning artificial heredity. A future physical implementation would require additional evidence and is outside the claims of this paper.

\section{Model}

\subsection{Germline representation and founders}
The germline is a normalized vector $G\in[-1,1]^{102}$. Ninety-nine loci use an autosomal channel and three use a dam-specific channel analogous only in computational role to a cytoplasmic contribution. The loci parameterize an $8\times8$ regulatory matrix $W$, decay constants $\lambda$, regulatory biases $b$, zygotic determinants $s_0$, morphogen sensitivities, diffusion, developed-network coupling, fate temperature, developmental duration, developmental noise, and morphogen steepness. Physical or biological interpretations are not implied by these names.

Four fixed founders A--D are generated from distinct rows of a Sylvester Hadamard construction, truncated to 102 coordinates and scaled to equal per-locus magnitude. This creates founders that are deliberately distinct but matched in norm. Because these are four fixed founders rather than a sample from a breeding population, all parental effects in this paper are fixed-design quantities. Accordingly, the present design establishes the existence and causal identifiability of biparental effects within this controlled founder geometry; it does not estimate their prevalence or average magnitude across an evolved or naturally sampled population of germlines.

\subsection{Recombination}
For each parent, a gametic vector is formed as
\begin{equation}
g=\operatorname{clip}(G+\xi,-1,1),\qquad \xi\sim\mathcal{N}(0,\sigma_{\mathrm{mut}}^2),
\end{equation}
with $\sigma_{\mathrm{mut}}=0.05$ in normalized coordinates. Each autosomal locus is then inherited from the dam or sire with probability $0.5$; the three dam-channel loci are inherited from the dam. The implementation stores a \emph{background} containing the recombination mask, both mutation vectors, and the developmental-noise seed. Holding this background fixed makes zygote formation deterministic given parental germlines and enables the H5 intervention.

\subsection{Development and differentiation}
The generic soma consists of eight initially equivalent modules located along a one-dimensional positional axis. Each module starts with zero regulatory state and runs the same zygotic network. For module $m$ and regulatory coordinate $i$,
\begin{align}
\frac{d z_i^{(m)}}{dt} &= -\lambda_i z_i^{(m)} + \tanh\left(\sum_j W_{ij}z_j^{(m)}+b_i+s_i^{(m)}\right)\\
&\quad +D_{\mathrm{diff}}\left(z_i^{(m-1)}+z_i^{(m+1)}-2z_i^{(m)}\right)+\sigma_{\mathrm{dev}}\,dW_t,
\end{align}
where positional input is
\begin{equation}
s_i^{(m)}=s_{0,i}+\mathrm{morph}_i\tanh(\mathrm{morph\_steep}\,p_m).
\end{equation}
Euler--Maruyama integration uses $dt=0.05$ and no-flux boundaries. The final regulatory state is frozen. A non-heritable, identical readout maps frozen states to module fates, oscillator parameters, time constants, biases, and pairwise couplings. The resulting developed network controls a differential-drive body abstraction in a fixed sensory field.

The model includes two no-development ablations. A structural ablation assigns the same $s_0$ state to every module; because all modules are identical, fate diversity collapses by construction. A more informative quasistatic ablation retains positional input while removing recurrence, diffusion, developmental noise, and developmental time.

\subsection{Newborn phenotype and absence of learning}
After development, the controller is fixed for a 200-time-unit evaluation. The six traits are mean speed, dominant gait frequency, signed turning bias, perturbation recovery time, oscillator coherence, and maximum exploration radius. Initial conditions and environmental parameters are identical for all newborns. No optimizer, reward-driven update, plasticity rule, gradient-based adaptation, or post-development parameter update is applied during newborn evaluation. Independent validation confirmed that the operational soma remains unchanged throughout this evaluation and that the evaluated phenotype therefore reflects the developed newborn state rather than post-birth adaptation.

Recovery time is censored at 90 time units if a perturbed trajectory does not return to a phase-invariant neighborhood of its nominal orbit. Censored individuals are retained. The diallel censoring rate for recovery is 26.9\%; recovery is therefore retained as the prespecified sixth outcome but interpreted as a secondary, censoring-limited measure and is not used as evidence for the headline heredity claims.

\section{Experimental Design and Validation}

Model constants and hypothesis decision rules were frozen after three disclosed non-degeneracy pilot corrections and before hypothesis statistics were computed. The corrections addressed a motor readout floor/ceiling artifact, insufficient fate-map dynamic range, and a phase-sensitive recovery definition. No model constant was subsequently changed. The computational runs preserve the complete configuration, random seed, software environment, and run-level provenance needed for reproducibility.

The main factorial experiment is a complete $4\times4$ dam-by-sire diallel with 40 offspring per cell ($n=640$) plus 160 parental-reference individuals. Reciprocal crosses are distinct. H1 is supported if at least one trait shows both parental main effects after Holm correction across six traits. H4 scores hybrid offspring as transgressive when they exceed clone-line means by more than two pooled within-line standard deviations and compares the hybrid and clone rates using a permutation test. The pre-specified analysis described A$\times$B as shorthand for the reciprocal hybrid comparison; the implemented H4 analysis pools A$\times$B and B$\times$A offspring, and the reported statistics and Figure~\ref{fig:transgression} follow that definition.

H5 uses 60 matched backgrounds. The base cross is A$\times$B; the alternative dam is C and the alternative sire is D. Within a background, the recombination mask, both mutation vectors, and developmental-noise seed are fixed. The intervention replaces one parental germline. For scale comparison, the design also includes same-parent \emph{stochastic-variation controls}: one re-draws mutation while keeping parent identity fixed (re-mutation) and one re-draws developmental noise (re-development). These controls are not ``causally unchanged'' states; they deliberately quantify phenotype displacement induced by stochastic variation without a parental-identity substitution.

Two no-development ablations, clone controls, a magnitude-matched random-germline control, and an ordinary direct-controller inheritance baseline are analyzed separately. Six three-generation lineages and a 20-point robustness sweep are exploratory/supplementary because the central claims do not require evolution across generations.

An independent numerical audit was completed before finalizing this manuscript. The fixed-effects ANOVA and PDVF calculations were recomputed directly from the archived simulation outputs and reproduced the reported values to machine precision. H5 uncertainty intervals were also recomputed with a paired bootstrap over matched backgrounds; the pre-specified paired sign-flip permutation tests and their statistical decisions were unchanged. The corresponding reproducibility materials are available with the public code and data release.

\section{Results}

\subsection{Primary result 1: both parental germlines structure newborn phenotype}

\begin{figure}[t]
\centering
\includegraphics[width=0.98\textwidth]{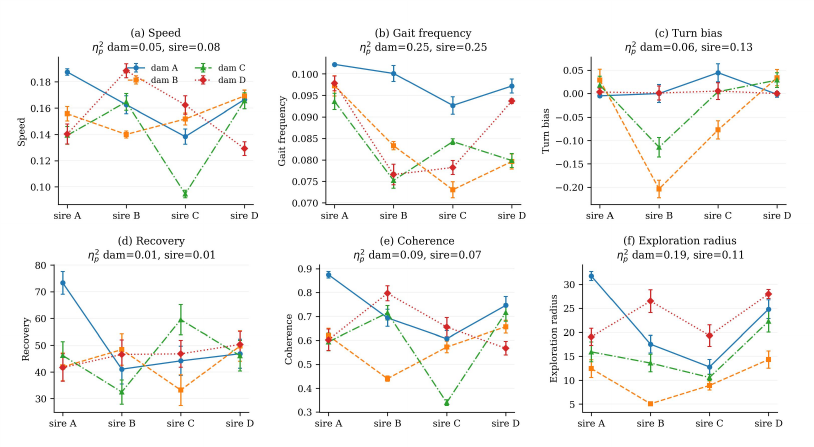}
\caption{Primary result 1: factorial parental contributions in the balanced $4\times4$ diallel ($40$ offspring per cell). Curves show cell means for each dam--sire combination; annotations report partial $\eta^2$ for dam and sire. The result demonstrates structured dependence of five newborn traits on both parental identities. Recovery is weak at the parental-main-effect level and is additionally affected by censoring.}
\label{fig:factorial}
\end{figure}

The $4\times4$ diallel shows pronounced factorial structure (Figure~\ref{fig:factorial}). Dam and sire main effects are significant after Holm correction for five of six traits: speed, gait frequency, turning bias, coherence, and exploration radius. Recovery time is the exception. Interaction terms are also substantial for several traits, emphasizing that this system should not be interpreted as a simple additive genetic model.

\begin{table}[t]
\centering
\caption{Factorial parental effects and parental developmental variance fraction (PDVF). Holm-adjusted permutation $p$ values are shown for dam and sire. PDVF is a design-specific variance-component fraction and is not narrow-sense heritability.}
\label{tab:h1}
\small
\begin{tabular}{lrrrrr}
\toprule
Trait & $\eta^2_p$ dam & $p_{\mathrm{dam}}$ & $\eta^2_p$ sire & $p_{\mathrm{sire}}$ & PDVF [95\% CI]\\
\midrule
Speed & .050 & .0012 & .077 & .0012 & .379 [.343,.459]\\
Gait frequency & .252 & .0012 & .252 & .0012 & .533 [.484,.602]\\
Turn bias & .059 & .0012 & .130 & .0012 & .360 [.303,.455]\\
Recovery time & .008 & .2026 & .009 & .1300 & .082 [.063,.178]\\
Coherence & .092 & .0012 & .072 & .0012 & .377 [.349,.449]\\
Exploration radius & .193 & .0012 & .106 & .0012 & .378 [.342,.457]\\
\bottomrule
\end{tabular}
\end{table}

The modeled parental variance fraction ranges from approximately .36 to .53 for the five principal traits, with gait frequency showing the strongest parental structure. Importantly, PDVF is not $h^2$. The founders are fixed, interactions are non-negligible, there is no additive breeding-population model, and ``environmental'' variance is not partitioned in the quantitative-genetic sense.

Reciprocal A$\times$B and B$\times$A crosses do not differ significantly on any of the six traits after correction, despite the model containing three dam-channel loci. Thus the architecture permits directional parent effects, but these particular founders do not yield statistically detectable reciprocal phenotypes.

\subsection{Primary result 2: matched germline substitution causes newborn phenotype shifts}

\begin{figure}[t]
\centering
\includegraphics[width=0.98\textwidth]{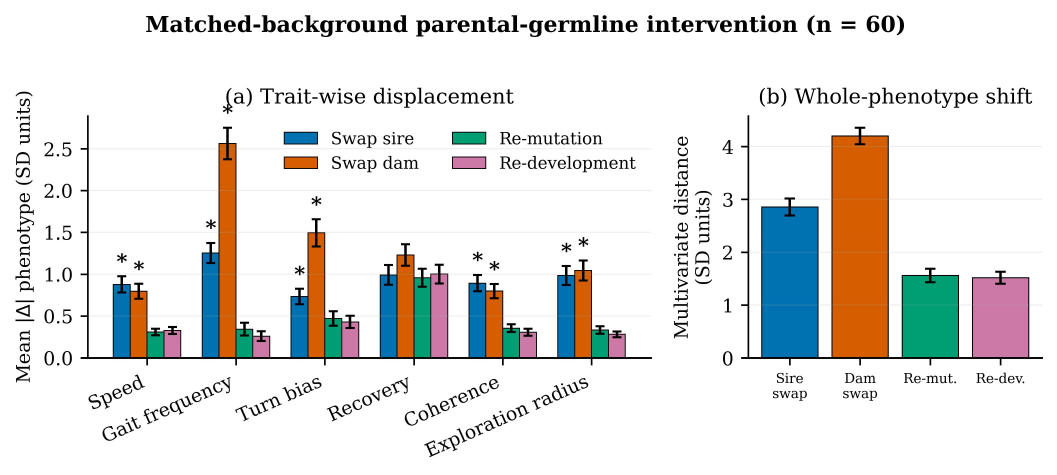}
\caption{Primary result 2: matched-background parental-germline interventions ($n=60$). Bars show mean absolute newborn phenotype displacement in standardized trait units. ``Re-mutation'' and ``re-development'' are stochastic-variation controls, not no-change null states. Significance markers denote contrasts that exceed the re-mutation control after Holm correction.}
\label{fig:swap}
\end{figure}

The diallel establishes factorial dependence, whereas the causal evidence is provided by H5. In H5, each A$\times$B newborn is compared with counterfactual siblings generated under the identical recombination mask, parental mutation vectors, and developmental-noise seed. Only the identity of one parental germline is substituted. Figure~\ref{fig:swap} shows that both sire and dam substitutions produce much larger changes than same-parent re-mutation or re-development for most traits.

Using a paired bootstrap appropriate to the matched design, sire substitution exceeds the re-mutation control by 0.571 standardized units for speed (95\% CI [0.386, 0.767]), 0.913 for gait [0.671, 1.178], 0.263 for turn bias [0.056, 0.466], 0.539 for coherence [0.341, 0.752], and 0.654 for exploration radius [0.426, 0.898]. The paired standardized effects $d_z$ for those contrasts are .75, .91, .32, .66, and .70, respectively. Recovery shows essentially no sire-substitution excess (0.034; CI spanning zero).

Dam substitution produces an excess of 0.489 for speed [0.296, 0.684], 2.222 for gait [1.836, 2.615], 1.023 for turning bias [0.711, 1.338], 0.444 for coherence [0.256, 0.634], and 0.713 for exploration radius [0.471, 0.966], with paired $d_z$ values of .63, 1.44, .82, .59, and .72. Recovery has a positive unadjusted paired-bootstrap interval, but its preregistered Holm-adjusted paired permutation $p=.0564$; it is therefore not classified as significant. The causal conclusion is consequently restricted to the five consistently supported traits.

Because the intervention changes one parental germline while preserving the remainder of the stochastic background, these results support a causal statement about the implemented computational architecture: parent identity is not merely correlated with newborn behavior; substituting a germline causes a phenotype shift under matched conditions.

\subsection{Secondary result: recombination generates transgressive offspring}

\begin{figure}[H]
\centering
\includegraphics[width=0.98\textwidth]{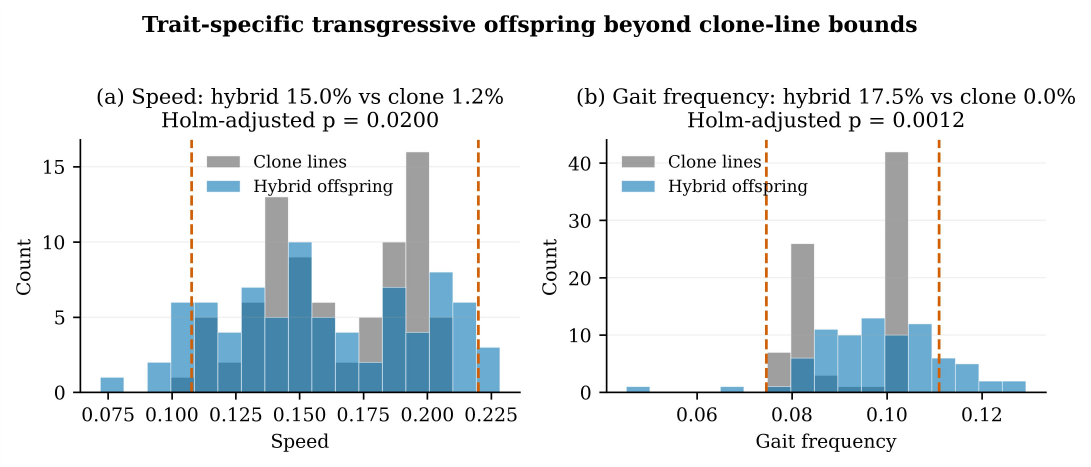}
\caption{Trait-specific transgressive offspring. Histograms compare the two parental clone lines (A$\times$A and B$\times$B) with reciprocal hybrid offspring (A$\times$B and B$\times$A). Dashed vertical lines mark the preregistered clone-line transgression bounds. Significant excess transgression is observed for speed and gait frequency after Holm correction.}
\label{fig:transgression}
\end{figure}

Under the prespecified transgression interval criterion, the implemented reciprocal-hybrid pool yields 15.0\% of A$\times$B/B$\times$A offspring that exceed the speed interval defined by the parental clone lines, versus 1.2\% of clone-line individuals; the Holm-adjusted permutation $p=.0200$. For gait frequency, 17.5\% of hybrids are transgressive versus 0\% of clone-line individuals ($p=.0012$). The criterion is not met after correction for turning bias, recovery, coherence, or exploration radius. Figure~\ref{fig:transgression} visualizes the two traits that survive correction. We therefore describe recombination novelty as trait-specific rather than a general property of every phenotype coordinate.

\subsection{The failed developmental-dependence hypothesis}

\begin{figure}[H]
\centering
\includegraphics[width=0.98\textwidth]{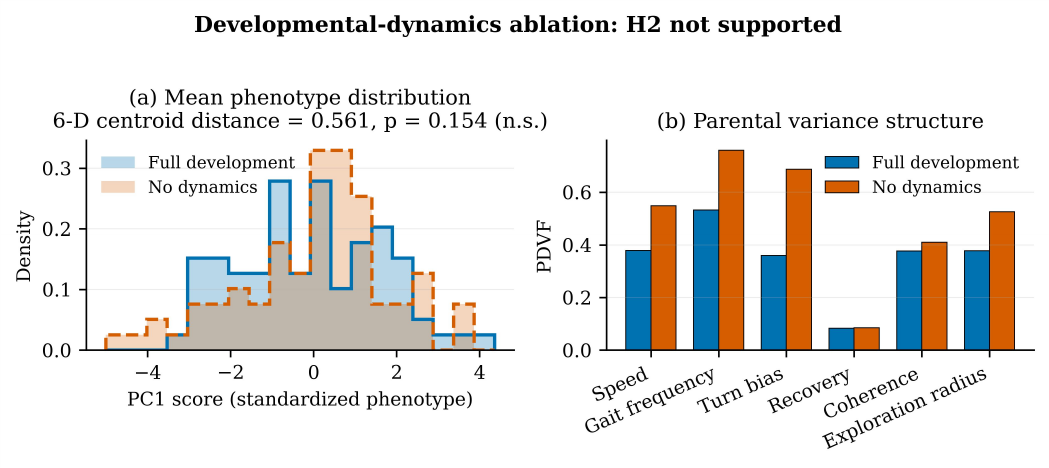}
\caption{Developmental-dynamics ablation. (a) A descriptive PC1 projection shows strong overlap between the full-development and quasistatic no-dynamics phenotypes; the reported $p=.154$ is the preregistered six-dimensional centroid permutation test, not a one-dimensional PC1 test. (b) PDVF changes under the quasistatic ablation, showing that recurrent dynamics alter parental variance structure even though the mean multivariate phenotype is not significantly displaced.}
\label{fig:h2ablation}
\end{figure}

H2 was preregistered as a conjunction: a no-development condition had to alter the multivariate phenotype distribution and the parental variance structure. The informative quasistatic ablation fails the first half. Its centroid is only 0.561 pooled-SD units from the full biparental condition, with permutation $p=.154$, and zero of six individual traits differs after correction. At the same time, parental variance fractions increase under the ablation for most traits (Figure~\ref{fig:h2ablation}). Thus H2 is \emph{not supported}.

This negative result constrains the mechanistic interpretation of Machine Zygote. In the present implementation, recurrent developmental dynamics are not required for the mean newborn phenotype to retain biparental structure. Instead, the dynamics and developmental noise reshape the parental variance structure and introduce additional non-parental variation. The present study therefore does not support the claim that development creates heredity; it supports the narrower conclusion that the germline--soma architecture expresses biparental pre-learning heredity while the tested recurrent dynamics modulate, rather than generate, that dependence.

\subsection{Robustness and exploratory generations}

\begin{figure}[H]
\centering
\includegraphics[width=0.98\textwidth]{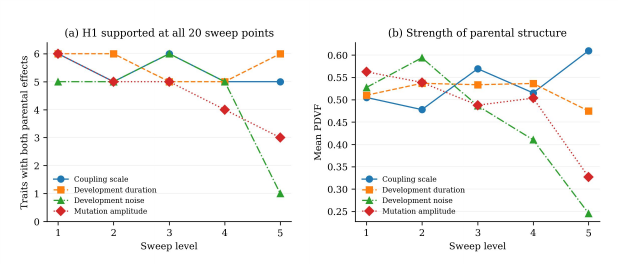}
\caption{Robustness sweep across coupling scale, developmental duration, developmental-noise scale, and mutation amplitude. Panel (a) reports the number of traits retaining significant effects from both parental channels; panel (b) summarizes the mean PDVF. The H1 criterion is recovered at all 20 tested parameter points.}
\label{fig:robustness}
\end{figure}

Across a supplementary sweep of coupling scale, developmental duration, developmental-noise scale, and mutation amplitude, the basic H1 criterion is recovered at all 20 tested parameter points (Figure~\ref{fig:robustness}). These sweeps use the parametric balanced-ANOVA $F$ tests for efficiency and should be viewed as robustness checks rather than independent replications of the preregistered permutation analysis.

The three-generation experiment contains no selection operator. It therefore demonstrates only recombination, mutation, developmental variation, and lineage drift. Although five of six traits show significant midparent--offspring slopes in the reported G2 individual-level regressions, family clustering makes those nominal $p$ values exploratory; we do not use them as evidence for the central claim. A cluster-aware multigeneration study would be preferable if transgenerational persistence becomes a major focus.

\section{Discussion}

\subsection{What the experiment establishes}

The principal evidential support comes from the convergence of a fixed-design factorial diallel and an individual-level matched-background intervention. In a balanced diallel, the identity of each parent explains structured variation in five newborn behavioral traits. In matched-background substitutions, replacing either parent's germline produces phenotype shifts that exceed stochastic re-mutation variation on the same five traits. All of these traits are measured before any learning mechanism is available. Together, these observations support \emph{causal biparental pre-learning heredity} as an operational property of the implemented simulation.

That phrase should be understood precisely. ``Biparental'' means that two separately represented parental germlines participate in zygote formation. ``Heredity'' means that parent-associated germline state systematically influences descendant phenotype. ``Pre-learning'' means that the phenotype is evaluated with a fixed controller before any adaptation. ``Causal'' refers specifically to the H5 substitution experiment, not to the fixed-effects diallel alone.

\subsection{Why this is not simply inherited controller weights}

The germline vector is not used as the newborn controller. It parameterizes a regulatory process and fixed differentiation map that construct the operational soma. The same generic soma substrate and readout are used for all individuals, and the direct-controller baseline is explicitly separated from the proposed pathway. This does not make the model biologically realistic, but it does distinguish its architecture from simply crossing final controller parameters.

At the same time, developmental and indirect encodings are well established \parencite{stanley2003,bongard2001}, and robot evolution already treats parental code as a source of offspring morphology and control \parencite{eiben2013,jelisavcic2019,luo2023}. The conceptual contribution therefore lies less in the existence of a genotype--phenotype map than in the experimental decomposition: parental identity, recombination, development, stochastic variation, direct-controller inheritance, and newborn learning are made separately manipulable variables.

\subsection{The failed H2 is informative}

H2 provides an important mechanistic boundary on the positive heredity findings. The initial hypothesis was that recurrent zygotic development would be necessary for inherited organization, but the quasistatic ablation does not support that claim. The positive H1 and H5 results therefore cannot be attributed specifically to recurrent temporal dynamics: the evidence supports causal germline--soma heredity without demonstrating that path-dependent ontogeny is necessary for the mean biparental phenotype. This result motivates a stronger criterion for future developmental models, in which mature organization must depend on temporal history in a way that cannot be reproduced by a matched static nonlinear mapping from zygotic parameters to final soma.

This suggests a sharper future test. A next-generation model should include an initially more weakly specified soma and require developmental history, local interaction, or perturbation recovery during ontogeny to determine final organization. The appropriate null would preserve all static positional information while removing temporal dynamics, as the quasistatic arm does here. A positive result would then demonstrate path-dependent development rather than merely an indirect encoding.

\subsection{Toward a physical Machine Zygote}

Nothing in the present simulation demonstrates physical heredity. The 102-dimensional germline is serialized software state, and the body is a differential-drive abstraction. A physically instantiated germline would require hereditary state to reside in material or device dynamics and to be physically transferred or recombined between parents and offspring. That would move the project toward the broader embodied-evolution and ontogenetic-hardware traditions \parencite{sipper1997,eiben2012,zykov2005}. Such an experiment is a future paper, not an interpretation of the current data.

A feasible progression is therefore: first establish computational causality, as done here; next implement a laboratory zygotic substrate whose components can be physically swapped while holding the remaining hardware fixed; and only then test whether inherited physical state survives into a pre-learning phenotype. This sequencing avoids presenting simulation as evidence of material heredity.

\section{Limitations}

First, the entire study is computational. No robot or material germline was built. Second, the four founder germlines were deliberately constructed and are not a random sample from any population. The reported PDVF is design-specific and should not be generalized as biological heritability; the fixed-founder design demonstrates effects within the tested geometry rather than population-level prevalence or average effect size. Third, the differentiation/readout rule is hand-designed, fixed, and non-heritable; this helps separate germline from soma but also constrains the class of phenotypes the model can express. Fourth, one body abstraction and one standardized sensory environment are used. Fifth, recovery time is censored for 26.9\% of diallel individuals. It is retained because it was a prespecified outcome, but it is treated as a secondary, censoring-limited measure and is not used to support the headline conclusions. Sixth, three model components were modified during disclosed non-degeneracy pilots before preregistered hypothesis analysis. Seventh, H1 permutation exchangeability is most naturally interpreted as testing factorial dependence under the model's balanced design; strong interaction means dam and sire effects are not purely additive. Eighth, the H5 intervals in this manuscript are independently verified paired-bootstrap intervals rather than the unpaired bootstrap intervals in the original analysis summary; the preregistered paired-permutation decisions are unchanged. Ninth, the multigeneration analysis lacks family-cluster-aware inference and is therefore exploratory. Finally, the term ``germline'' describes computational role, not biological composition.

\section{Conclusion}

Machine Zygote provides a reproducible computational test bed for a narrow but foundational question: can an artificial newborn carry causal contributions from two parental hereditary channels before it learns? In this implementation, the answer is yes for five of six measured traits. A balanced diallel detects structured dependence on both parental identities, and matched-background parental substitutions produce substantial newborn phenotype shifts beyond stochastic re-mutation variation. Recombination yields transgressive speed and gait phenotypes. However, a preregistered hypothesis that recurrent developmental dynamics are necessary is not supported; development modulates the variance structure rather than determining the mean biparental phenotype in this implementation. This separates the supported heredity claim from the stronger, unsupported claim of developmental necessity.

Accordingly, the present findings do not establish physical machine heredity or the invention of artificial embryogeny. Their value is methodological and architectural: the framework makes biparental heredity, newborn phenotype, development, learning, and causal parental intervention simultaneously explicit and experimentally separable. That framework provides a stricter starting point for future physical Machine Zygote systems and for subsequent studies of transgenerational memory and the evolution of developmental processes.

\section*{Data, Code, and Reproducibility}
The complete code and data release is publicly available at \href{https://github.com/LyesSaadSaoud/machine-zygote}{Machine Zygote public repository}. It contains the computational model, frozen experimental configuration, raw and processed simulation outputs, analysis materials, validation records, and figure-generation resources required to reproduce the reported study. The complete experiment suite contains 6076 simulated individuals. The supplementary material documents the numerical audit and the additional analyses supporting the main results.

\section*{Transparency on Generative-AI Assistance}
Generative AI tools were used to assist with English-language refinement and to improve the clarity and readability of the manuscript.

\printbibliography

\clearpage
\newgeometry{margin=0.85in}
\singlespacing
\small
\setcounter{section}{0}
\setcounter{subsection}{0}
\setcounter{figure}{0}
\setcounter{table}{0}
\renewcommand{\thesection}{S\arabic{section}}
\renewcommand{\thesubsection}{S\arabic{section}.\arabic{subsection}}
\renewcommand{\thefigure}{S\arabic{figure}}
\renewcommand{\thetable}{S\arabic{table}}
\renewcommand{\theHsection}{S\arabic{section}}
\renewcommand{\theHsubsection}{S\arabic{section}.\arabic{subsection}}
\renewcommand{\theHfigure}{S\arabic{figure}}
\renewcommand{\theHtable}{S\arabic{table}}
\begin{center}
{\Large\bfseries Supplementary Material}\\[0.5em]
{\large Machine Zygote: Causal Biparental Heredity Before Learning in a Germline--Soma Artificial Agent}\\[0.5em]
Lyes Saad Saoud\\
Independent Researcher\\
Chicago, Illinois, USA; Abu Dhabi, UAE
\end{center}
\vspace{0.8em}

\section{Purpose and scope}
This supplement contains material useful for scientific audit but not needed in the main narrative: preregistration details, implementation specifics, complete sample-size and statistical tables, secondary and negative results, and exploratory analyses. Detailed software documentation is maintained separately in the public reproducibility repository at \href{https://github.com/LyesSaadSaoud/machine-zygote}{Machine Zygote public repository}.

The study is a \textbf{computational proof of concept only}. No physical robot, physical germline, or material zygote was built or measured. ``Germline,'' ``zygote,'' ``dam,'' and ``sire'' denote computational roles in the model rather than biological composition.

\section{Pre-specified Analysis Plan and Validation}

The public reproducibility package contains the frozen analysis plan prepared after the non-degeneracy pilots and before hypothesis statistics were computed, together with the archived simulation outputs and run-level provenance. The master seed is \texttt{20260823} and the configuration digest is \texttt{8270e321ac1fc1c7}.

\subsection{Disclosed pilot changes}
Three changes occurred before hypothesis testing:
\begin{enumerate}
\item \textbf{Motor readout.} A sigmoidal wheel readout made mean speed nearly constant. It was replaced by a rectified mapping so a silent motor pool produces no forward drive.
\item \textbf{Fate-map dynamic range.} Fate logits were mean-subtracted across modules and multiplied by a fixed gain of 5.0 because preliminary states otherwise produced nearly uniform fate weights.
\item \textbf{Recovery metric.} Recovery was changed from raw twin-state convergence to a transverse, phase-invariant distance from the nominal limit-cycle orbit because an impulse produces a persistent phase shift even after dynamical recovery.
\end{enumerate}
No subsequent model changes were made before confirmatory hypothesis testing. These pilot modifications are disclosed because they define the finalized model evaluated in the reported analyses.

\subsection{Preregistered hypotheses}
\begin{description}
\item[H1: Biparental contribution.] Supported if at least one common trait has significant dam and sire effects after Holm correction in the balanced diallel.
\item[H2: Developmental dependence.] Supported only if removing recurrent developmental dynamics changes both the multivariate phenotype distribution and at least one component of the parental variance structure. The primary comparison uses the quasistatic ablation that preserves positional input while removing temporal dynamics.
\item[H3: Pre-learning inheritance.] Supported if H1 holds and the newborn phenotype is measured under a fixed evaluation protocol with no learning or adaptive parameter update.
\item[H4: Recombination novelty.] Supported if the A$\times$B transgression rate significantly exceeds the clone-line rate against the same preregistered threshold.
\item[H5: Causal swap.] Supported if replacing a parental germline under matched stochastic background produces a paired phenotype displacement greater than same-parent re-mutation variation for at least one trait after Holm correction.
\end{description}

\begin{table}[htbp]
\centering
\caption{Preregistered outcomes.}
\begin{tabular}{lll}
\toprule
Hypothesis & Outcome & Main interpretation\\
\midrule
H1 & Supported & Both parental identities structure newborn phenotype\\
H2 & Not supported & Dynamics not necessary for mean biparental phenotype\\
H3 & Supported & H1 effects occur before learning\\
H4 & Supported & Trait-specific transgressive recombination\\
H5 & Supported & Matched parental substitution changes phenotype\\
\bottomrule
\end{tabular}
\end{table}

\section{Complete computational model}

\subsection{Germline}
Each individual has a germline $G\in[-1,1]^{102}$. Denormalization of a normalized coordinate $u$ to a physical model parameter with range $[l,h]$ is
\begin{equation}
p=l+\frac{u+1}{2}(h-l).
\end{equation}
The 102 loci are partitioned as follows.

\begin{longtable}{@{}p{0.14\textwidth}r p{0.16\textwidth} p{0.13\textwidth} p{0.28\textwidth}@{}}
\caption{Germline blocks and model roles.}\\
\toprule
Block & Size & Range & Channel & Role\\
\midrule
\endfirsthead
\toprule
Block & Size & Range & Channel & Role\\
\midrule
\endhead
$W$ & 64 & $[-0.50,0.50]$ & autosomal & $8\times8$ regulatory coupling matrix\\
$\lambda$ & 8 & $[0.50,2.00]$ & autosomal & gene-product decay\\
$b$ & 8 & $[-0.60,0.60]$ & autosomal & constitutive bias\\
$s_0$ & 8 & $[-0.80,0.80]$ & autosomal & zygotic determinant\\
$morph$ & 8 & $[-1,1]$ & autosomal & positional-morphogen sensitivity\\
$D_{diff}$ & 1 & $[0,0.50]$ & autosomal & inter-module diffusion\\
$\kappa_J$ & 1 & $[0.15,1]$ & autosomal & developed coupling gain\\
$fate\_temp$ & 1 & $[0.50,2]$ & autosomal & fate softmax temperature\\
$T_{dev}$ & 1 & $[12,36]$ & dam channel & development duration\\
$\sigma_{dev}$ & 1 & $[0.010,0.060]$ & dam channel & developmental noise\\
$morph\_steep$ & 1 & $[0.60,2.50]$ & dam channel & morphogen steepness\\
\bottomrule
\end{longtable}

Founders A--D are generated by taking rows 1--4 of a Sylvester Hadamard construction, truncating them to 102 coordinates, and multiplying by 0.7. This gives equal coordinate magnitudes and similar norms without post-hoc tuning of one founder to be stronger than another. The magnitude-matched random-germline control draws each coordinate as $\pm0.7$ with equal probability. These founders are deliberately constructed rather than sampled from an evolved population. The resulting diallel therefore supports an existence and causal-identification claim within this controlled founder geometry, not an estimate of population-level prevalence or average effect size across arbitrary germlines.

\subsection{Zygote and reproduction}
Independent gametes are
\begin{equation}
g=\operatorname{clip}(G+\xi,-1,1),\qquad \xi\sim\mathcal{N}(0,0.05^2).
\end{equation}
For 99 autosomal coordinates, a recombination mask chooses dam or sire with probability $0.5$ at each coordinate. The three dam-channel loci are always inherited from the dam. Clone crosses therefore still have within-line variation because their two gametes receive independent mutation draws.

For matched interventions, the stochastic background comprises the autosomal recombination mask, both mutation vectors, and the developmental-noise seed. Holding this background fixed makes the child zygote a deterministic function of the two input germlines and isolates the effect of substituting one parental germline.

\subsection{Development}
Eight soma modules begin with identical zero regulatory states at positions $p_m\in[-1,1]$. Positional drive is
\begin{equation}
s_i^{(m)}=s_{0,i}+morph_i\tanh(morph\_steep\,p_m).
\end{equation}
The regulatory dynamics are
\begin{align}
\frac{dz_i^{(m)}}{dt} =&-\lambda_i z_i^{(m)}+\tanh\!\left(\sum_jW_{ij}z_j^{(m)}+b_i+s_i^{(m)}\right)\\
&+D_{diff}\left(z_i^{(m-1)}+z_i^{(m+1)}-2z_i^{(m)}\right)+\sigma_{dev}\,dW_t.
\end{align}
Euler--Maruyama integration uses $dt=0.05$, no-flux boundaries, and a zygote-specific developmental duration. At the end of development, the regulatory state is frozen.

\subsection{Fixed differentiation and soma readout}
The readout from frozen developmental state to operational soma is identical in every individual and not inherited. The first five regulatory dimensions yield fate logits, the remaining dimensions affect oscillator/coupling parameters. Fate logits are mean-centered across modules and scaled by a fixed gain of 5 before a softmax. Developed pairwise coupling is proportional to similarity between frozen expression profiles, scaled by an inherited global gain. This construction intentionally makes the germline different from a final controller: connectivity is computed from developed state rather than stored directly as an inherited matrix.

\subsection{Newborn dynamics}
The operational network follows
\begin{align}
\tau_u\dot u_m &= -u_m+w_m r_m+\sum_{n\neq m}J_{mn}r_n+bias_m+I_m-a_m,\\
\tau_{a,m}\dot a_m &= g_mr_m-a_m,\qquad r_m=\tanh(u_m).
\end{align}
A position-dependent sensory input drives the network. Developed motor pools generate rectified left/right wheel commands, with standard differential-drive kinematics
\begin{equation}
\dot x=v\cos\theta,\quad \dot y=v\sin\theta,\quad \dot\theta=\omega.
\end{equation}
All newborns share the same initial state and environment.

\subsection{Phenotype traits}
The six reported traits are: mean speed, dominant gait frequency, signed turning bias, perturbation recovery time, amplitude-weighted oscillator coherence, and maximum exploration radius. Recovery uses a phase-invariant transverse distance to the nominal orbit; failures to recover are censored at 90 and retained. In the diallel, 26.9\% of recovery observations are censored, whereas motionless, nonoscillatory, and nonfinite flags are each 0\%. Because of this censoring, recovery is retained as the prespecified sixth outcome but interpreted as a secondary, censoring-limited measure rather than evidence for the headline heredity claims.

\section{Absence of learning: verification}
The no-learning condition was verified directly against the released implementation. Automated validation confirmed that the newborn soma is unchanged by evaluation, that no learning or parameter-update operation is invoked on the evaluation path, and that all individuals begin from identical evaluation initial conditions. These checks support the interpretation of the measured phenotype as a pre-learning newborn phenotype.

\section{Experimental sample sizes}
\begin{table}[htbp]
\centering
\caption{Experiment sizes.}
\begin{tabular}{lll}
\toprule
Experiment & Design & Size\\
\midrule
Diallel & $4\times4$ dam $\times$ sire & 640 offspring + 160 references\\
Causal swap & 60 backgrounds $\times$ 5 arms & 300 evaluations\\
Baselines & 7 conditions & 80 per condition where applicable\\
Lineages & 6 three-generation lineages & 576 individuals\\
Robustness & 4 knobs $\times$ 5 levels & 30 replicates/cell\\
Total & complete suite & 6076 simulated individuals\\
\bottomrule
\end{tabular}
\end{table}

\section{Statistical procedures}
The diallel is balanced by construction. Fixed-effects two-way ANOVA partitions dam, sire, interaction, and residual sums of squares. Reported quantities include $F$, partial $\eta^2$, $\omega^2$, and effect estimates. The preregistered main-effect permutation procedure permutes one parent label within levels of the other; because interactions are present, the safest interpretation is a test of factorial dependence under the balanced design rather than a population-genetic additive effect.

Holm--Bonferroni adjustment is applied across the six traits within each hypothesis family. Bootstrap and permutation procedures use 5000 resamples or permutations, respectively. The PDVF variance-component summary is
\begin{align}
V_d &= \max\left(\frac{\mathrm{MS}_d-\mathrm{MS}_e}{n\,b},0\right),\\
V_s &= \max\left(\frac{\mathrm{MS}_s-\mathrm{MS}_e}{n\,a},0\right),\\
V_{ds} &= \max\left(\frac{\mathrm{MS}_{ds}-\mathrm{MS}_e}{n},0\right),\\
\mathrm{PDVF} &= \frac{V_d+V_s+V_{ds}}{V_d+V_s+V_{ds}+\mathrm{MS}_e}.
\end{align}
Here $a=b=4$ are the numbers of dam and sire founder identities and $n=40$ is the number of offspring per diallel cell. PDVF is a design-specific summary for this fixed founder geometry and is not narrow-sense heritability.

For H5, the original analysis correctly uses paired sign-flip permutation tests but used an unpaired bootstrap for the reported difference-of-means intervals. The independent reanalysis addresses this mismatch by resampling matched rows and bootstrapping the paired difference directly. Statistical decisions remain based on the original preregistered paired permutation tests; the corrected intervals improve uncertainty reporting without changing the verdicts.

\section{Independent Numerical Verification}
The central H1 fixed-effects quantities were independently recomputed from the archived raw diallel outputs. The maximum absolute discrepancy between the recomputation and the reported summary for the $F$ statistics and component PDVF values was zero at floating-point precision.

\begin{table}[htbp]
\centering
\caption{Independent H1 re-computation.}
\small
\begin{tabular}{lrrrr}
\toprule
Trait & $F_{dam}$ & $F_{sire}$ & $F_{int}$ & PDVF\\
\midrule
Speed & 11.030 & 17.250 & 18.862 & .379\\
Gait & 70.089 & 69.953 & 12.066 & .533\\
Turn bias & 13.138 & 30.979 & 12.952 & .360\\
Recovery & 1.586 & 1.908 & 4.220 & .082\\
Coherence & 21.121 & 16.245 & 16.313 & .377\\
Exploration & 49.869 & 24.692 & 7.165 & .378\\
\bottomrule
\end{tabular}
\end{table}

\subsection{Corrected paired H5 uncertainty intervals}
\begin{longtable}{llrrrr}
\caption{Matched germline-swap effects relative to the same-parent re-mutation control. $d_z$ is the paired standardized mean difference. The confidence intervals are post-audit paired-bootstrap intervals; hypothesis decisions use the preregistered Holm-adjusted paired permutation tests.}\\
\toprule
Swap & Trait & Mean shift & Excess & 95\% CI & $d_z$\\
\midrule
\endfirsthead
\toprule
Swap & Trait & Mean shift & Excess & 95\% CI & $d_z$\\
\midrule
\endhead
Sire & Speed & .879 & .571 & [.386,.767] & .748\\
Sire & Gait & 1.255 & .913 & [.671,1.178] & .906\\
Sire & Turn & .733 & .263 & [.056,.466] & .322\\
Sire & Recovery & .992 & .034 & [-.257,.321] & .030\\
Sire & Coherence & .894 & .539 & [.341,.752] & .659\\
Sire & Exploration & .986 & .654 & [.426,.898] & .698\\
Dam & Speed & .796 & .489 & [.296,.684] & .628\\
Dam & Gait & 2.564 & 2.222 & [1.836,2.615] & 1.439\\
Dam & Turn & 1.494 & 1.023 & [.711,1.338] & .818\\
Dam & Recovery & 1.231 & .273 & [.003,.539] & .254\\
Dam & Coherence & .798 & .444 & [.256,.634] & .589\\
Dam & Exploration & 1.045 & .713 & [.471,.966] & .721\\
\bottomrule
\end{longtable}
Although the unadjusted paired-bootstrap interval for dam recovery is positive, its preregistered Holm-adjusted paired-permutation $p=.0564$, so it is not called significant.

\section{Secondary result: developmental ablations (H2)}

\begin{figure}[htbp]
\centering
\includegraphics[width=0.96\textwidth]{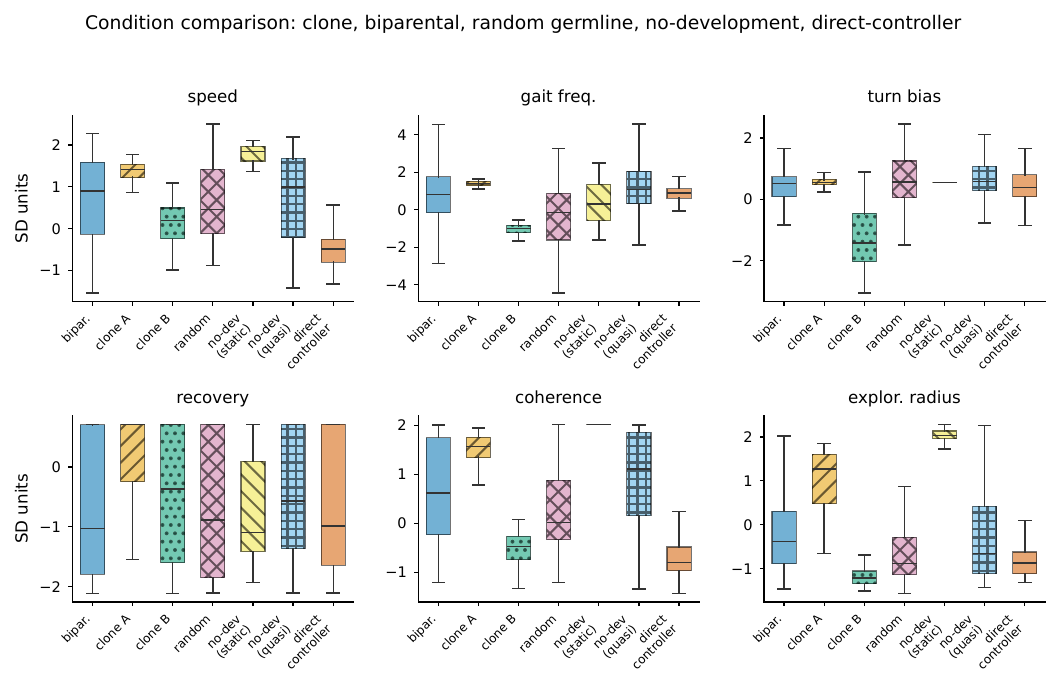}
\caption{Baseline and ablation comparisons. The informative quasistatic control removes recurrent developmental dynamics; the static control is partly degenerate because identical modules cannot differentiate.}
\end{figure}

The quasistatic no-dynamics condition has multivariate centroid distance 0.561 pooled-SD units from the full biparental model with permutation $p=.154$ and zero of six significant traitwise differences after correction. H2 therefore fails its preregistered conjunctive criterion. Nevertheless, PDVF rises under the ablation for speed (.379 $\rightarrow$ .548), gait (.533 $\rightarrow$ .760), turn (.360 $\rightarrow$ .688), recovery (.082 $\rightarrow$ .085), coherence (.377 $\rightarrow$ .410), and exploration (.378 $\rightarrow$ .525). The dynamics add non-parental variability even though they do not create the mean biparental phenotype in this implementation. Accordingly, the supported H1/H5 heredity findings should not be interpreted as evidence that recurrent temporal dynamics are themselves causally necessary for the mean biparental phenotype; that stronger mechanistic claim remains unsupported in this model.

\section{Secondary result: phenotype-space structure}
\begin{figure}[htbp]
\centering
\includegraphics[width=0.92\textwidth]{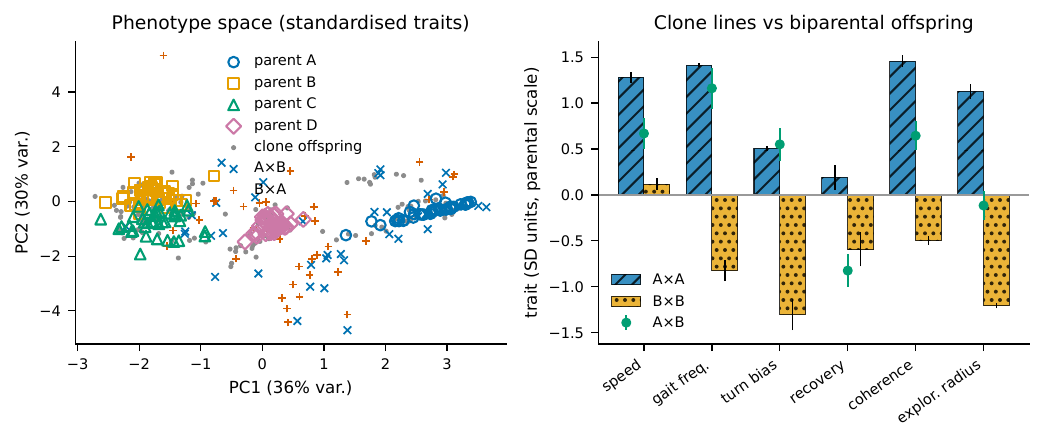}
\caption{Parent and offspring phenotype-space projection. The visualization is descriptive and is not used as primary causal evidence.}
\end{figure}

\section{Secondary result: development trajectories}
\begin{figure}[htbp]
\centering
\includegraphics[width=0.94\textwidth]{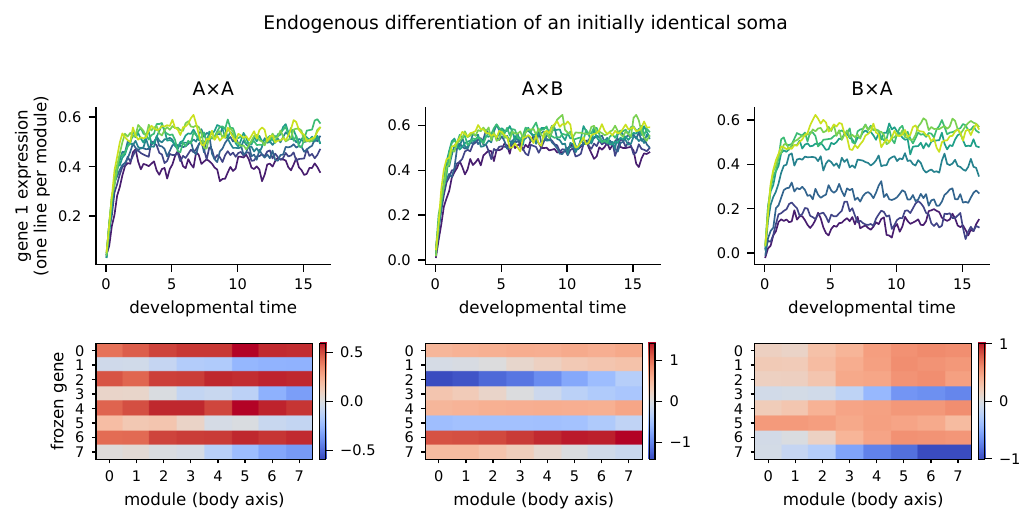}
\caption{Representative developmental trajectories and frozen soma state. This figure illustrates the implemented regulatory dynamics; H2 shows that such dynamics are not necessary for the mean parental phenotype under the tested quasistatic ablation.}
\end{figure}

\section{Secondary result: transgressive recombination (H4)}
A trait is preregistered as transgressive when the offspring lies outside the interval between the two parental clone-line means expanded by two pooled within-line standard deviations. The same interval is applied to clone individuals to estimate a stochastic false-transgression rate.

\begin{table}[htbp]
\centering
\caption{Preregistered H4 transgression test.}
\small
\begin{tabular}{lrrr}
\toprule
Trait & Hybrid rate & Clone rate & Holm $p$\\
\midrule
Speed & 15.0\% & 1.2\% & .0200\\
Gait & 17.5\% & 0.0\% & .0012\\
Turn bias & 11.2\% & 2.5\% & .2368\\
Recovery & 0.0\% & 0.0\% & 1.0000\\
Coherence & 0.0\% & 0.0\% & 1.0000\\
Exploration & 6.2\% & 0.0\% & .2368\\
\bottomrule
\end{tabular}
\end{table}
The result supports trait-specific transgressive recombination for speed and gait, not a global novelty claim.

\section{Reciprocal crosses}
The model includes three dam-channel loci, so reciprocal effects are possible. In A$\times$B versus B$\times$A, however, no trait is significant after Holm correction. The mean differences are small relative to uncertainty for speed, gait, turning, and recovery. Exploration has a raw confidence interval just above zero but remains nonsignificant after family correction. This null result should be preserved because it limits claims of parent-role asymmetry.

\section{Multigeneration experiment: exploratory only}
\begin{figure}[htbp]
\centering
\includegraphics[width=0.94\textwidth]{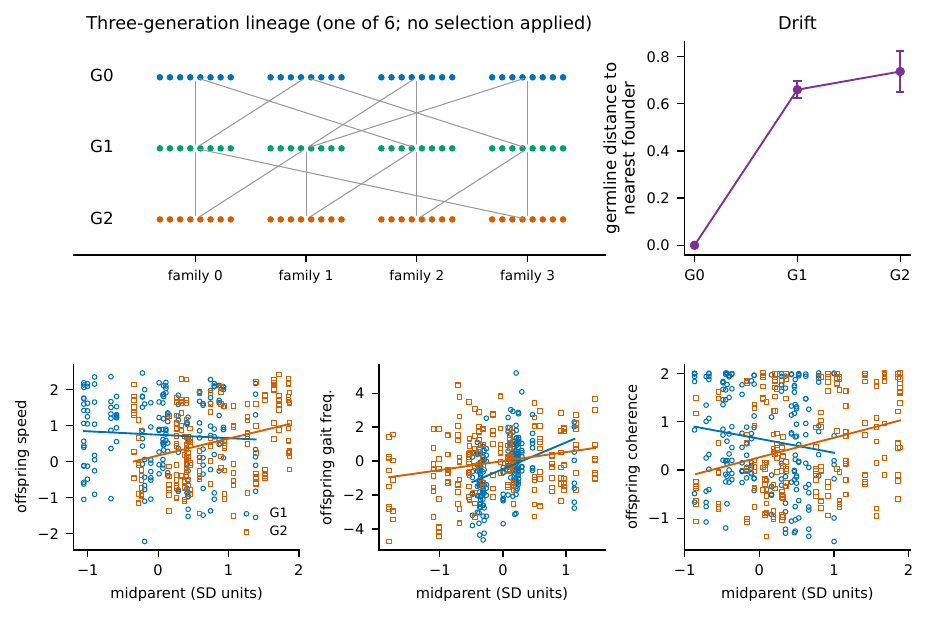}
\caption{Three-generation simulation with no selection. Changes reflect recombination, mutation, developmental stochasticity, and drift, not evolution.}
\end{figure}

The original individual-level analysis reports G2 midparent--offspring slopes significant for five of six traits. However, individuals are nested within families/lineages and share midparent values. The main manuscript therefore does not treat these nominal $p$ values as confirmatory evidence. A future transgenerational study should use family/lineage as the independent unit, a mixed model, or a cluster bootstrap.

\section{Robustness sweep}
\begin{figure}[htbp]
\centering
\includegraphics[width=0.96\textwidth]{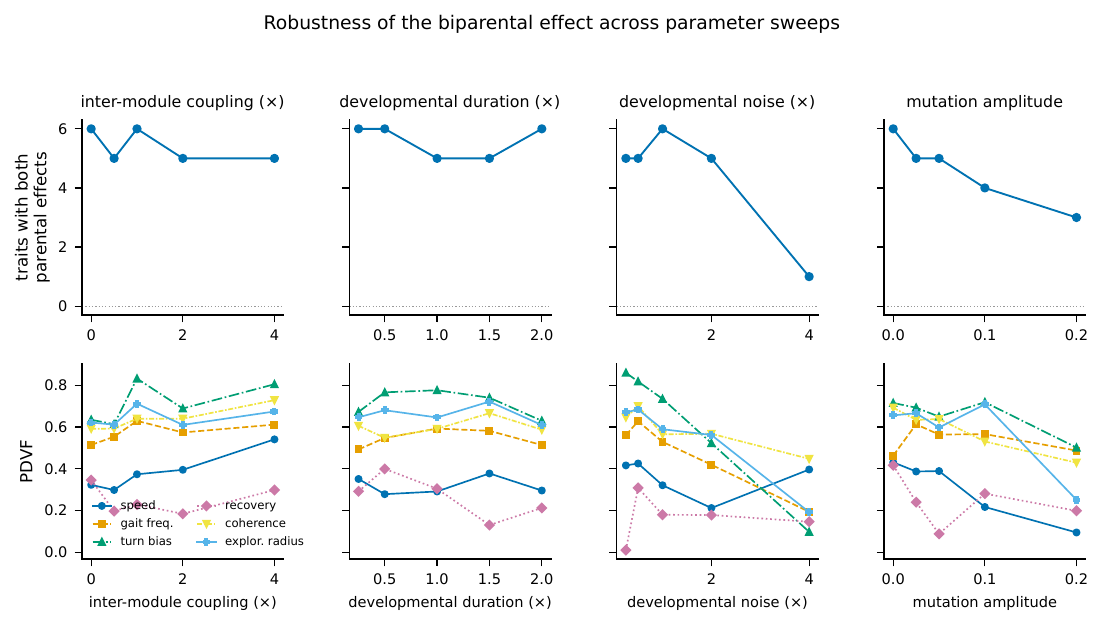}
\caption{Robustness sweep across coupling scale, developmental duration, developmental noise, and mutation amplitude. The H1 criterion is detected at all 20 tested sweep points.}
\end{figure}

The sweep tests coupling scales $0,0.5,1,2,4$; developmental-duration multipliers $0.25,0.5,1,1.5,2$; developmental-noise multipliers $0.25,0.5,1,2,4$; and mutation standard deviations $0,0.025,0.05,0.10,0.20$. Each point has 30 replicates per tested cross setting. For computational tractability this experiment uses the parametric balanced-ANOVA $F$ tests rather than the 5000-permutation main analysis, so it is a sensitivity check rather than a second preregistered confirmatory experiment.

\section{Public repository and reproducibility}
The complete reproducibility package is publicly available at \href{https://github.com/LyesSaadSaoud/machine-zygote}{Machine Zygote public repository}. It contains the computational model, frozen experimental configuration and analysis plan, raw and processed simulation outputs, provenance records, numerical-audit material, validation resources, and figure-generation materials. Detailed software organization and execution instructions are maintained in the repository so that this supplement can remain focused on scientific methods, analyses, and results.

For archival identification, the reported experiment suite uses master seed \texttt{20260823}, configuration digest \texttt{8270e321ac1fc1c7}, and 6076 simulated individuals. The supplementary figures were regenerated from the same archived numerical outputs; only presentation attributes such as plotting colors were changed for readability, with no alteration to the underlying values or statistical calculations.
\end{document}